\documentclass[information,article,accept,pdftex,moreauthors]{Definitions/mdpi}
\firstpage{1}
\pubvolume{16}
\issuenum{3}
\articlenumber{203}
\pubyear{2025}
\copyrightyear{2025}
\externaleditor{Gholamreza Anbarjafari (Shahab)} 
\datereceived{21 January 2025}
\daterevised{13 February 2025} 
\dateaccepted{24 February 2025}
\datepublished{5 March 2025}
\hreflink{https://\linebreak  doi.org/10.3390/info16030203} 

\Title{\textls[-20]{Attention Mechanism-Based Cognition-Level Scene Understanding}}

\TitleCitation{Attention Mechanism-Based Cognition-Level Scene Understanding}

\Author{{Xuejiao} 
Tang $^{1}$ and {Wenbin Zhang}  $^{2,}$*}

\AuthorNames{Xuejiao Tang and Wenbin Zhang}

\AuthorCitation{{Tang, X.;} 
Zhang, W.}

\address{%
$^{1}$ \quad {Institute for Information Processing, }Leibniz {University} 
Hannover, {Welfengarten 1, \linebreak  30167 Hannover, Germany;} 
xuejiao2020@gmail.com\\
$^{2}$ \quad {Knight Foundation School of Computing \& Information Sciences,} 
Florida International {University}, \linebreak  {Miami, FL~33199, USA} 

}

\corres{Correspondence:  wenbin.zhang@fiu.edu}

\abstract{Given a question--image input, a visual commonsense reasoning (VCR) model predicts an answer with a corresponding rationale, which requires inference abilities based on real-world knowledge. The VCR task, which calls for exploiting multi-source information as well as learning different levels of understanding and extensive commonsense
knowledge, is a cognition-level scene understanding challenge. The VCR task has aroused researchers' interests due to its wide range of applications, including visual question answering, automated vehicle systems, and clinical decision support. Previous approaches to solving the VCR task have generally relied on pre-training
or exploiting memory with long-term dependency relationship-encoded models.
However, these approaches suffer from a lack of generalizability and a loss of information in long sequences. In this work, we propose a parallel attention-based cognitive VCR network, termed PAVCR, which fuses visual--textual information efficiently and encodes semantic information in parallel to enable the model to capture rich information for cognition-level inference. Extensive experiments show that the proposed model yields significant improvements over existing methods on the benchmark VCR dataset. Moreover, the proposed model provides an intuitive interpretation of visual commonsense reasoning.}

\keyword{visual commonsense reasoning; visual understanding}

\begin{document}




\section{Introduction}\label{sec1}

Visual understanding is an important research domain with a long history that attracts extensive models such as Mask RCNN~\cite{vuola2019mask}, ResNet~\cite{he2016deep,andreas2016neural}, and UNet~\cite{barkau1996unet}. These models are successfully employed in various visual understanding tasks such as action recognition, image classification, pose estimation, and~visual search~\cite{DBLP:journals/corr/PapandreouZKTTB17}. Most gain high-level understanding by identifying the objects in view based on visual input~\cite{sriram1986applications,zhang2023pmc}. However, reliable visual scene understanding requires both recognition- and cognition-level visual understanding and their seamless integration. More specifically, it is desirable to identify the objects of interest to infer their actions, intents, and~mental states to have a comprehensive and reliable understanding of the visual input. While this complex task comes naturally to humans, existing visual understanding systems lack the capacity for higher-order cognition inference~\cite{zellers2019recognition}.

Recent research in visual understanding has shifted inference from recognition-level to cognition-level, which contains more complex relationship inferences to improve cognition-level visual understanding~\cite{henderson2016end}. This directly leads to four significant directions for cognition-level visual understanding research: (1) image generation~\cite{gregor2015draw,kingma2014adam}, which aims to generate images from a given text description; (2) image caption~\cite{vinyals2015show}, which focuses on generating text description from given images; (3) visual question answering~\cite{patro2018differential,yu2017multi1,liu2023q2atransformer,hegde2023making}, whose goal is predicting correct answers for given image-and-question pairs; and (4) visual commonsense reasoning (VCR)~\cite{zellers2019recognition}, which additionally provides rational explanations along with question answering, and has gained considerable attention~\cite{yu2020ernie,wu2017visual,sun2019ernie}.

The VCR task is a challenging mountain to climb. The~model must  {develop a} 
human-level inference ability to solve a cognition-level visual task~\cite{lu2023multiscale}. While this might be easy for humans, since we have a reserve of knowledge and excellent, innate reasoning abilities, it is challenging for state-of-the-art AI systems. Recently, a~growing body of \mbox{research~\cite{zellers2019recognition,yu2020ernie,ben2017mutan,DBLP:journals/corr/abs-1910-14671,lu2019vilbert,Malinowski_2015,maslan2015one,kumar2016ask}} has aimed to solve this formidable task. In~their noteworthy attempts towards achieving VCR, these works typically necessitate pre-training on large-scale data before performing VCR tasks. Therefore, they usually fit well on properties the pre-training data possess, but~their generalization in other tasks is far from guaranteed~\cite{chen2019uniter}. To~remove this dependence on pre-training, another line of research focuses on directly learning the architecture of a system to find straightforward solutions for VCR~\cite{DBLP:journals/corr/abs-1910-14671}. However, these methods suffer commonsense-information loss where the commonsense reasoning ability of the model is limited. Further, successful VCR would also require efficient multimodal fusion to fuse visual and textual information together, which remains a daunting task~\cite{wu2018multi,natarajan2012multimodal}.

To address the aforementioned challenges, we propose a parallel-structure-based model, PAVCR, which encodes commonsense between sentences with lower information loss. We first design a multimodal fusion layer to fuse visual and textual \mbox{information~\cite{baltruvsaitis2018multimodal,sharma-etal-2018-conceptual,lu2016hierarchical,ilievski2016focused,yan2022spca}.} Then, we introduce a commonsense encoder to enhance the inference ability of the proposed model. Finally, we present a commonsense encoder to enhance the inference ability of our proposed model.
More specifically, the~contributions of this research~are as follows:
\begin{enumerate}
\item We theoretically analyze the inadequacy of previous work~\cite{tang2021cognitive} and introduce a new, more effective model to reduce the sequential computation. In~\cite{tang2021cognitive}, the~computational complexity of related signals from two positions grows with the distance between the positions, which results in difficulties in learning dependencies among positions for a sequential task. Our proposed multimodal feature fusion layer and commonsense encoder submodules have a novel parallel attention structure to limit the number of~operations.
\item {We use an}
original, innovative model for the VCR task, which represents cognition-level scene understanding, leading to significant improvements in performance.
\item {We use a}
novel multimodal fusion layer that fuses visual and textual information, resulting in considerable improvements in results on VCR tasks.
\item {We use a} 
new commonsense encoder layer with a parallel-structure attention mechanism and memory cell that avoids information loss while storing the extracted knowledge and extracted commonsense between queries and responses.
\item {We conduct} 
thorough comparisons with popular methods for VCR tasks and ablation studies via extensive experiments on real-world data. Finally, we present a deeper analysis of the results via a case study and qualitative analysis.
\end{enumerate}

This work is an extension of our previous work~\cite{tang2021interpretable}. The~significant changes include the following: (i) a newly proposed multimodal fusion layer for visual--textual information fusion, (ii) a parallel-structure-based commonsense encoder, which enables the model to capture more information without long dependency, along with a memory cell for commonsense storage, (iii) a case study to discuss the superior ability of the proposed model, and,~finally, (iv) an ablation study to validate the proposed multimodal fusion layer and commonsense encoder~layer.

The remainder of this paper is structured as follows. Related work on question answering (QA), especially VCR, is reviewed in Section~\ref{chap:relatedwork}. Section~\ref{chap:npf} succinctly covers notation. We present our proposed model in Section~\ref{chap:framework} and detail how it works to handle VCR tasks. The data, metrics, and~experiments are described in Section~\ref{chap:er}, and~Section~\ref{cs} compares our proposed model to the base model. Finally, quantitative results from PAVCR are presented in Section~\ref{qe}, and~we conclude with Section~\ref{chap:conclusion}.





\section{Related~Work}
\label{chap:relatedwork}

From individual object-level scene understanding~\cite{vuola2019mask,antol2015vqa,srivastava2012multimodal,krishna2017visual,dey2017gate,lu2016visual,zhang2023reducing}, which aims at object instance segmentation and image recognition, to~visual relationship detection~\cite{sabes1995advances,xiong2016dynamic,yang2019auto,zaremba2014recurrent,huang2019attention,wolfe2015visual}, which captures the relationship between any two objects in images or videos, state-of-the-art visual-understanding models have achieved remarkable progress~\cite{DBLP:journals/corr/CarreiraZ17,lugood}. However, there is still a long way to go, since an ideal visual system must have the ability to grasp the deeper meaning behind a scene.
To aid in that, recent research on visual understanding has shifted inference from the recognition level to the cognition level, which contains more complex relationship inferences.
Rowan~et~al.~\cite{zellers2019recognition} further formulated visual commonsense reasoning as the VCR task, an~important step towards reliable visual understanding, and~benchmarked the VCR dataset. Specifically, the~VCR dataset is sampled from a large pool of movie clips in which most scenes refer to logic inferences. For~example, ``Why isn't Tom sitting next to David?'' requires high-order inference ability about the scene to select the correct answer from available choices. Work towards achieving VCR generally falls into one of the following two categories based on whether a pre-training dataset is necessary or~not.

The first line of research, pre-training approaches, trains the model on a large-scale dataset and then fine-tunes the model for downstream tasks. Recent works include ERNIE-ViL-large~\cite{yu2020ernie} and UNITER-large~\cite{chen2019uniter}. While the former learns semantic relationship understanding for scene graph prediction, the~latter is pre-trained to learn joint image{--}text 
representations. However, the~generalizability of these models relies heavily on the pre-training dataset and therefore is not~guaranteed.

The other line of research focuses on studying the model structure to find a straightforward solution for the VCR task.
It focuses on encoding the relationship between sentences using sequence-to-sequence-based encoding methods. These methods infer rationales by encoding the long dependency relationship between sentences (see, e.g.,~R2C~\cite{zellers2019recognition}, TAB-VCR~\cite{DBLP:journals/corr/abs-1910-14671}, DMVCR~\cite{tang2021cognitive}, RevisitedVQA~\cite{jabri2016revisiting}, BottomUpTopDown~\cite{anderson2018bottom}, MLB~\cite{kim2016hadamard}, MUTAN~\cite{ben2017mutan}).
However, these models face significant reasoning information loss due to the long dependency structure, and~it is hard for them to infer reason based on commonsense reasoning about the world.
More recently, CAN~\cite{tang2021interpretable} proposed a co-attention network to ease model training and enhance the ability to capture relationships between sentences and semantic information from surrounding words, which has led to a remarkable improvement in the VCR task.
Our work resembles this method, which is independent of a large-scale pre-training dataset. Two distinctions in our proposed work are the following: (i) a cross-attention-based multimodal unit designed to fuse visual information from images and textual information from sentences (i.e., from~language), and a
(ii) a parallel-structure co-attention network with a memory cell for storing commonsense rather than a long-dependency structure network to enhance the capability of capturing semantic information in the case of long~sentences.

In addition, recently, the task of visual commonsense generation has seen significant advancements. Models like VisualCOMET~\cite{park2020visualcomet}, KM BART~\cite{xing2021km}, CE-BART~\cite{kim2022bart}, and~Dynamic Debiasing Network~\cite{kim2023dynamic} have extended commonsense reasoning beyond static visual scenes to incorporate temporal dynamics. VisualCOMET~\cite{park2020visualcomet}, for~example, not only reasons about what is happening in a still image, but also predicts the events that occurred before and after the scene, and infers the intentions of the individuals in the image. This dynamic context reasoning is an important step towards advancing visual commonsense, as~it considers the temporal flow and human intent in visual scenes. KM BART~\cite{xing2021km} and CE-BART~\cite{kim2022bart} further build on this by leveraging knowledge-enhanced models and cause-and-effect reasoning, enhancing their ability to infer relationships and predict future events in a scene. These works demonstrate the growing importance of dynamic commonsense generation and the ability to reason about the context and sequence of events in visual scenes. While our approach primarily focuses on still images, the~insights from these dynamic models are invaluable for future work that aims to combine temporal reasoning with visual commonsense tasks. Despite these advancements, commonsense reasoning remains a central challenge in visual commonsense tasks for several reasons as follows. (1) The Complexity of Real-World Situations: Real-world scenarios often require nuanced reasoning that goes beyond surface-level visual patterns. Commonsense reasoning helps bridge the gap between visual information and human-like understanding, making it essential for tasks like {VQA}~\cite{yu2019deep,wang2018fvqa,andreas2016learning,shrestha2019answer,lin2014microsoft} and commonsense generation. (2) Contextual Understanding: Visual scenes are often ambiguous and require an understanding of context to interpret them correctly. Commonsense knowledge allows the model to infer relationships between objects and events that are not explicitly present in the image but are part of human experience and intuition. (3) Practical Applications: Commonsense reasoning is critical for applications across various domains, from~autonomous systems to robotics and human--computer interaction. These systems require models that can reason effectively and make decisions that align with human understandings of the~world.

Recent advancements in multimodal integration for high-level tasks have focused on cross-attention methods to combine visual and textual information effectively. One foundational work, on multimodal Factorized Bilinear (MFB) Pooling~\cite{yu2017multi2}, efficiently combines visual and textual features for visual question answering (VQA), addressing the high computational complexity of traditional bilinear pooling. It integrates a co-attention mechanism to jointly learn image and question attention, enhancing fine-grained representation. The~combined MFB and co-attention model achieve state-of-the-art performance in VQA tasks while improving computational efficiency. ViLT~\cite{kim2021vilt} proposed a minimal vision-and-language pre-training model that simplifies visual input processing by eliminating convolutions, making it significantly faster than previous models while maintaining competitive or improved performance on downstream tasks. Flamingo~\cite{alayrac2022flamingo} is a Visual Language model that excels at few-shot learning, rapidly adapting to various image and video tasks by leveraging large-scale multimodal training and architectural innovations for seamless integration of visual and textual data. Multilevel Attention networks~\cite{yu2017multi1} address challenges by integrating semantic attention for high-level concepts and visual attention for spatial region inference. Bidirectional Attention Flow for Machine Comprehension~\cite{seo2018bidirectionalattentionflowmachine} uses a multistage hierarchical process with bidirectional attention to generate query-aware context representations without early summarization, achieving state-of-the-art results. Although~these methods contribute significantly to multimodal fusion, they still face challenges with long-dependency~structures.

\nocite{zhang2024inpractice, zhang2024fairness, zhang2019faht, zhang2019fairness, zhang2020feat, zhang2020flexible, zhang2020online, zhang2020learning, zhang2021farf, zhang2021fair, zhang2022longitudinal, zhang2023censored, zhang2025fairness, zhang2023fairness, wang2016wearable, zhang2017hybrid, zhang2017phd, zhang2016using, zhang2018content, tang2020using, zhang2014comparison, zhang2020deep, wang2021harmonic, liu2022improving, zhang2021autoencoder, zhang2018deterministic, tang2019internet, tang2020data, zhang2021disentangled, huang2021lstm, liu2021research, liu2023segdroid, guyet2022incremental, zhang2021conditional, cai2023exploring}

\section{Notations and Problem~Formulation}
\label{chap:npf}
In this section, we succinctly describe necessary notation, the~format of the dataset typically used for the VCR task, and how it may be employed to test for the different subtasks that comprise~VCR.

\begin{figure}[H]

\begin{adjustwidth}{-\extralength}{0cm}
\centering 
\includegraphics[width=1.3\textwidth]{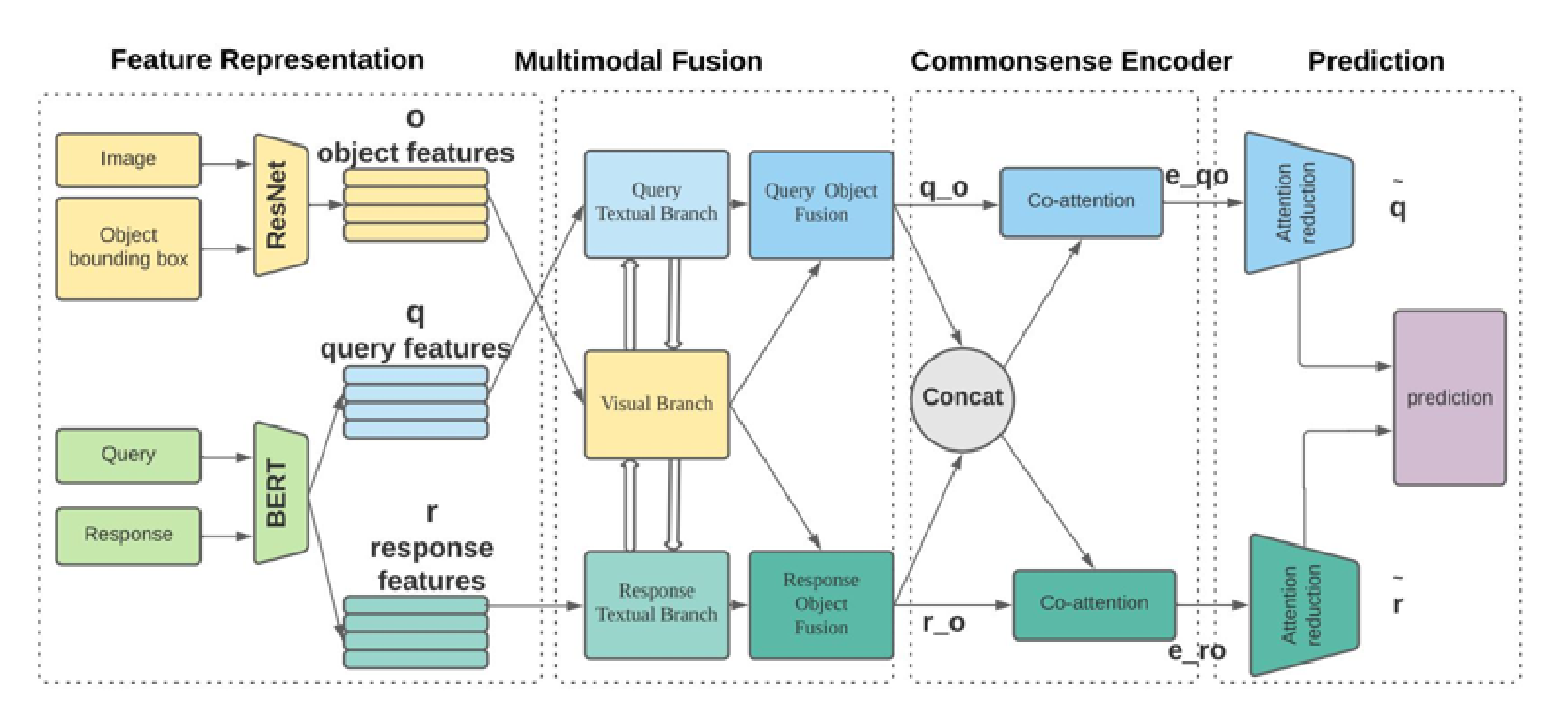}
\end{adjustwidth}
\caption{\label{fig:framework}Proposed~framework.}
\end{figure}

The VCR dataset~\cite{zellers2019recognition} consists of 290 k{-}
labeled subsets. Each subset is composed of an image with one to three associated questions. Each question in turn has four candidate answers and four candidate rationales, with~one correct answer and rationale. The~overarching task is formulated as a combination of three subtasks: (1) predicting the correct answer for a given question and image ($Q \rightarrow A$); (2) predicting the correct rationale for a given question, image, and~correct answer ($QA \rightarrow R$); and (3) predicting the correct answer and rationale for a given image and question ($Q \rightarrow AR$). Additionally, we defined two language inputs, query $q \in \{q_1,q_2,\cdots,q_n\}$ and response $r \in \{r_1,r_2,\cdots,r_n\}$, as~reflected in  {Figure}
~\ref{fig:framework}.
\vspace{-12pt}

\begin{figure}[H]
\includegraphics[width=0.9\textwidth]{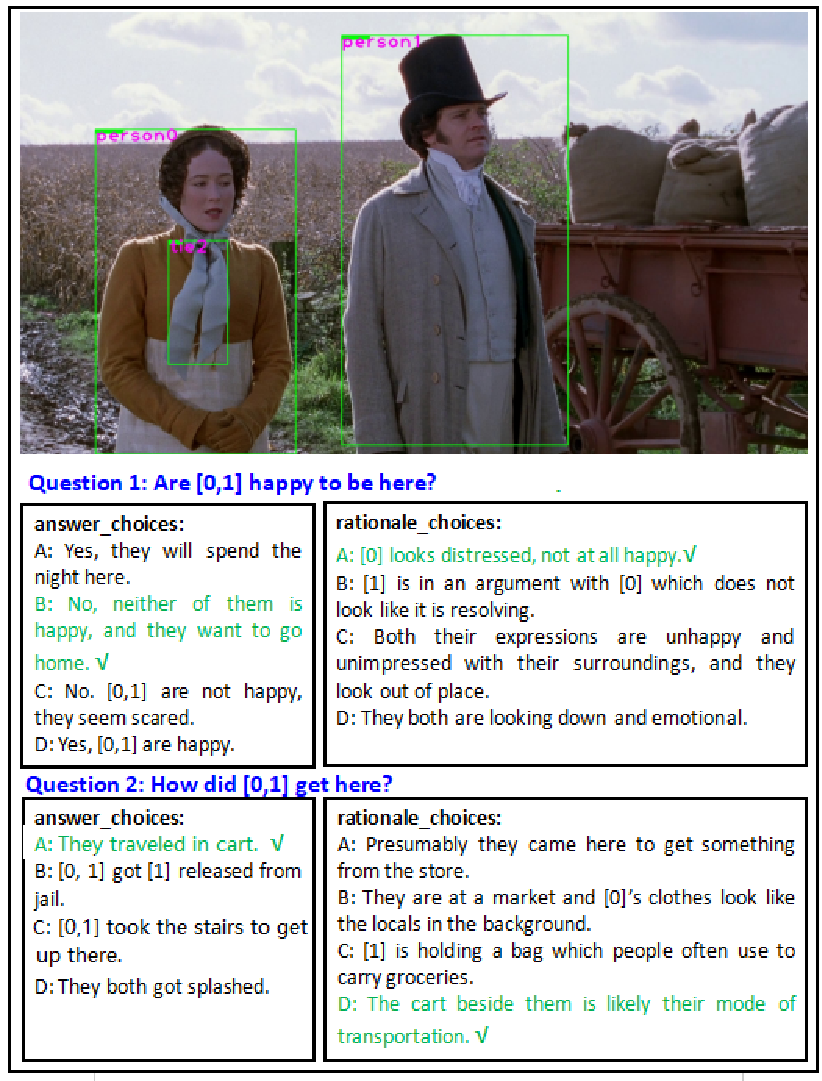}
\caption{An example of VCR running.}
\label{fig:running examp}
\end{figure}

In the $Q \rightarrow A$ subtask, query $q$ is the question, and~response $r$ is the answer choice.
In the $QA \rightarrow R$ subtask, query $q$ becomes the question together with the correct answer, while the possible rationales constitute the response $r$.
For example, in~Figure~\ref{fig:running examp}, the~$Q \rightarrow A$ subtask in Question 1 predicts the correct answer choice for a given image and question. Here, $q$ is the given question (``Are [0,1] happy to be here?''), and~$r$ is the given answer choices (``A: Yes, they will $\cdots$”, ``B: No, neither of them is happy,$\cdots$'', ``C: No, [0,1] took the stairs $\cdots$'', ``D: They both $\cdots$''). Compared to $Q \rightarrow A$ subtask, the~$QA \rightarrow R$ subtask predicts the rationale for the given image, question, and~correct answer. Here, $q$ is the question (``Are [0,1] happy to be here?'') along with the correct answer (``B: No, neither of them is happy, and~they want to go home.''), and $r$ is the rationale choices (``A: [0] looks distressed $\cdots$'', ``B: [1] is in an argument with [0] $\cdots$'', ``C: Both their expression $\cdots$'', ``D: They both $\cdots$'').

\section{Proposed~Framework}
\label{chap:framework}
In this section, we delve into our proposed framework in depth. Figure~\ref{fig:framework} illustrates our proposed framework's learning process. It consists of four layers: a feature representation layer, a~multimodal fusion layer, a~commonsense encoder layer, and~a prediction layer. The~first layer captures language and image features and converts them into dense representations. The~representations are then fed into the multimodal fusion layer to generate meaningful contexts of language-image fused information~\cite{lee2020generating,noh2016training,kim2016multimodal,le2019application,He_2019_ICCV,chen2019uniter}. Next, these fused features are fed into a commonsense encoder layer consisting of a co-attention unit and a memory cell to support commonsense storage. Finally, a~prediction layer predicts the correct answer or rationale. We explore each layer in~detail.

\subsection{Feature Representation~Layer}
\label{frl}


Extracting informative features from multi-source information plays a vital role in any machine learning application, especially in our context, where the features themselves are learning targets. As~shown in Figure~\ref{fig:framework}, the~inputs for the image feature extraction are the image along with its objects. The~objects are specified using bounding boxes, which serve as reference points for objects within an image. These bounding boxes are processed through RoiAlign to ensure precise alignment with the feature maps. For~the feature extraction process, the~deep network ResNet50~\cite{you2018imagenet} backbone is used, with~fine tuning applied to the final block of the network after RoiAlign. During~fine tuning, the~learning rate is set to  \linebreak  {1 $\times$ 10$^{-4}$,} 
and~the optimizer used is Adam, with~$\beta_1 = 0.9$, $\beta_2 = 0.999$, and~\linebreak  epsilon =  {1 $\times$ 10$^{-8}$.} 
For~data augmentation, we apply random horizontal flipping, scaling, and~color jittering to improve the robustness of the model and prevent overfitting. Additionally, we resize the input images to a consistent size of $224 \times 224$ pixels before feeding them into the~network.

In terms of architectural choices, skip connections~\cite{he2016deep} are utilized to mitigate the vanishing gradient problem during training. While skip connections help in preserving gradients across layers, we considered other methods like batch normalization and residual scaling. However, after~evaluating these techniques, we found that skip connections were more effective for the particular structure of our model, as~they allowed for smoother gradient flow through the network during the fine-tuning phase. The~decision to use skip connections was driven by their ability to improve convergence speed without introducing additional computational overhead compared to batch normalization or residual~scaling.

\textbf{{Language embedding.} 
}
Language embeddings are obtained by transforming raw input sentences into low-dimensional embeddings. The~embeddings are extracted using an attention mechanism with a parallel structure~\cite{devlin2018bert}. The~query represented by $q \in \{q_1,q_2,\cdots,q_n\}$ refers to a question in the question-answering task ($Q \rightarrow A$), and~a question paired with its correct answer in the reasoning task ($QA \rightarrow R$). Responses $r\{r_1,r_2,\cdots,r_n\}$ refer to answer candidates in the question-answering task ($Q \rightarrow A$), and~rationale candidates in the reasoning task ($QA \rightarrow R$).
Note that the sentences contain tags related to objects in the image. For~example, see Figure~\ref{fig:running examp} and the question ``Are [0,1] happy to be here?'' The [0,1] are tags set to identify objects in the image (i.e., the~object features of person 1 and person 2).

\textbf{Object embedding.} The images are sourced from movie clips. To~ensure a selection of images with rich information, a~filter is used to select images containing more than two objects each~\cite{zellers2019recognition}. These object features are then extracted with a residual connected deep network~\cite{he2016deep}. The~deep network outputs object features with low-dimensional embeddings $o \in \{o_1,o_2,\cdots,o_n\}$.

\subsection{Multimodal Feature Fusion~Layer}
\label{chap:mffl}

\begin{figure}[H]
\includegraphics[width=0.95\textwidth]{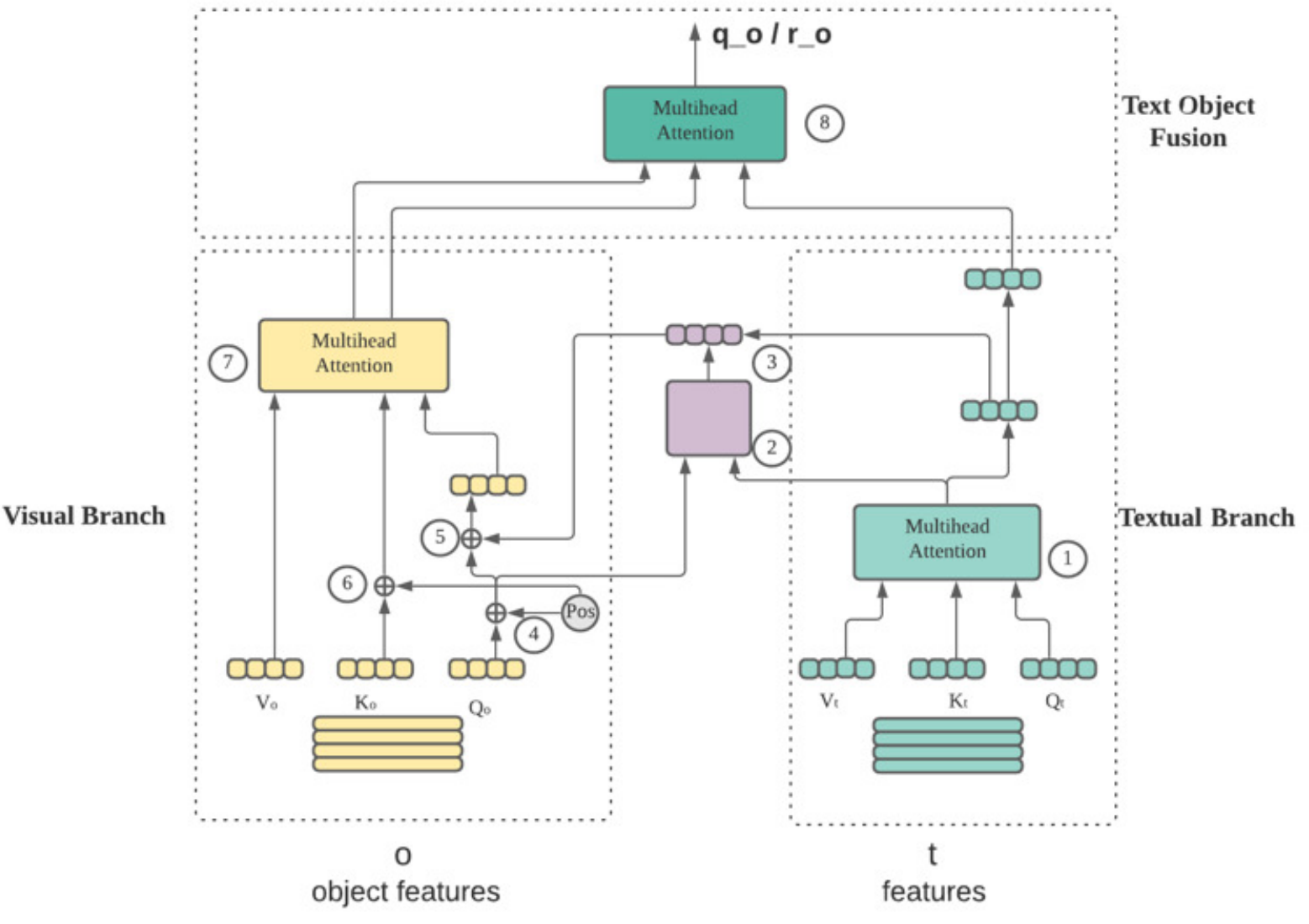}
\caption{{Multimodal} 
Feature Fusion~Layer.}
\label{fig:gr}
\end{figure}

Figure~\ref{fig:gr} illustrates the multimodal feature fusion layer. This layer aims to learn fused visual--textual features with semantic, discriminative visual features under the guidance of the textual description without harming their location ability~\cite{du2021visual}. These learned fused features enable the model to learn the ability to capture the context-level semantics of both vision and text. The~multimodal feature fusion layer has the following structure: (1)~a text branch to supply text features regarding a query and its responses with attention information for the related object features from the image; (2) a visual branch unit to supply mixed visual--textual information; (3) a text object fusion unit to enable the capability of the model in capturing the context-level semantics of both visual and textual information. Each unit is described in detail~below.

\textbf{Textual Branch Unit.} The textual branch unit is designed to supply textual information while extracting semantic information from words. To~this end, previously extracted textual features from queries or responses are regarded as value $V_t$, query $Q_t$, and~key $K_t$.
We employ multi-head attention~\cite{vaswani2017attention,sukhbaatar2015end} to obtain intended information from surrounding words. Formally  {put,} 
\begin{equation}
\label{1}
X_1 = V_tsoftmax(\frac{Q_tK_t}{\sqrt{d}})
\end{equation}
\noindent where $V_t$, $Q_t$, and~$K_t$ are the embedded textual features from queries or responses, and~$d$ is the dimension of the embeddings. (See operation \textcircled{1} in Figure~\ref{fig:gr}.)

The output, $X_1$, can supply rich textual information for visual--textual~fusion.

\textbf{Visual Branch Unit.} The visual branch is designed to learn joint visual--textual features, which fuses textual information into visual features. Specifically, the~previously extracted object features $O$ are considered query $Q_v$, keys $K_v$, and~values $V_v$ for the visual branch's self-attention computation. Position encodings (``Pos'' in Figure~\ref{fig:gr}) are added before weighted attention computations to identify the order of the objects and the words. Mathematically,
\begin{equation}
PE_{pos,2i} = sin(\frac{pos}{\frac{10000^{2i}}{d_{model}}}),
PE_{pos,2i+1} = cos(\frac{pos}{\frac{10000^{2i}}{d_{model}}})
\end{equation}
\noindent where $PE$ represents position encoding, $pos$ is the word position in sentence, $i$ is dimension, and~$d_{model}$ is the embedding~dimension.

Visual sequential order is crucial for semantic information because it could lead to incorrect meanings in a sentence. Therefore, the~visual branch unit first takes the previously extracted object features $O$ as the input of position encoding. The~operation is illuminated in Figure~\ref{fig:gr} in \textcircled{4} and \textcircled{6}, and~can be formulated as follows:
\begin{equation}
Q_{ve} = K_{ve} = norm(O^i)+PE(O^i)
\end{equation}

Next, the~output is fed into $softmax$ (peration \textcircled{2} in Figure~\ref{fig:gr}) to obtain a weighted sum expressed as follows:
\begin{equation}
\label{2}
X_2 = softmax(\frac{Q_{ve}X_1}{\sqrt{d}})
\end{equation}
\noindent where $X_1$ is textual information containing weighted attention information of each word and comes from operation \textcircled{1} in the textual~unit.

Recall that the aim of the visual branch unit is to learn joint visual--textual representation. To~this end, operation \textcircled{5} concatenates visual--textual information from the output of operation \textcircled{3}, which can be defined as
\begin{equation}
\label{5}
Q_{ve} \oplus X_3
\end{equation}
\noindent where $\oplus$ represents the concatenate operation and $X_3$ is the output of operation \textcircled{3}, which can be represented as
\begin{equation}
\label{3}
X_3 = X_2X_1
\end{equation}

Next, a~multi-head attention in operation \textcircled{7} is deployed to obtain fused visual representations, which contain weighted attention information from text~features.

\textbf{Text--Object Fusion Unit.}
After the visual branch unit, the~model has learned rich visual information containing weighted textual information and semantic information from surrounding words. To~obtain context-level semantics regarding both textual and visual information, we leverage another multi-head attention operation to learn text--object~representations.

Finally, the~multimodal fusion layer outputs the fused visual and textual information in the form of $q_o$ and $r_o$ (Figure~\ref{fig:gr}). $q_o$ represents the fused query and object features, while $r_o$ denotes the fused response and object~features.

\subsection{Commonsense~Encoder}
\label{ce}
\begin{figure}[H]

\begin{adjustwidth}{-\extralength}{0cm}
\centering 
\includegraphics[width=1.15\textwidth]{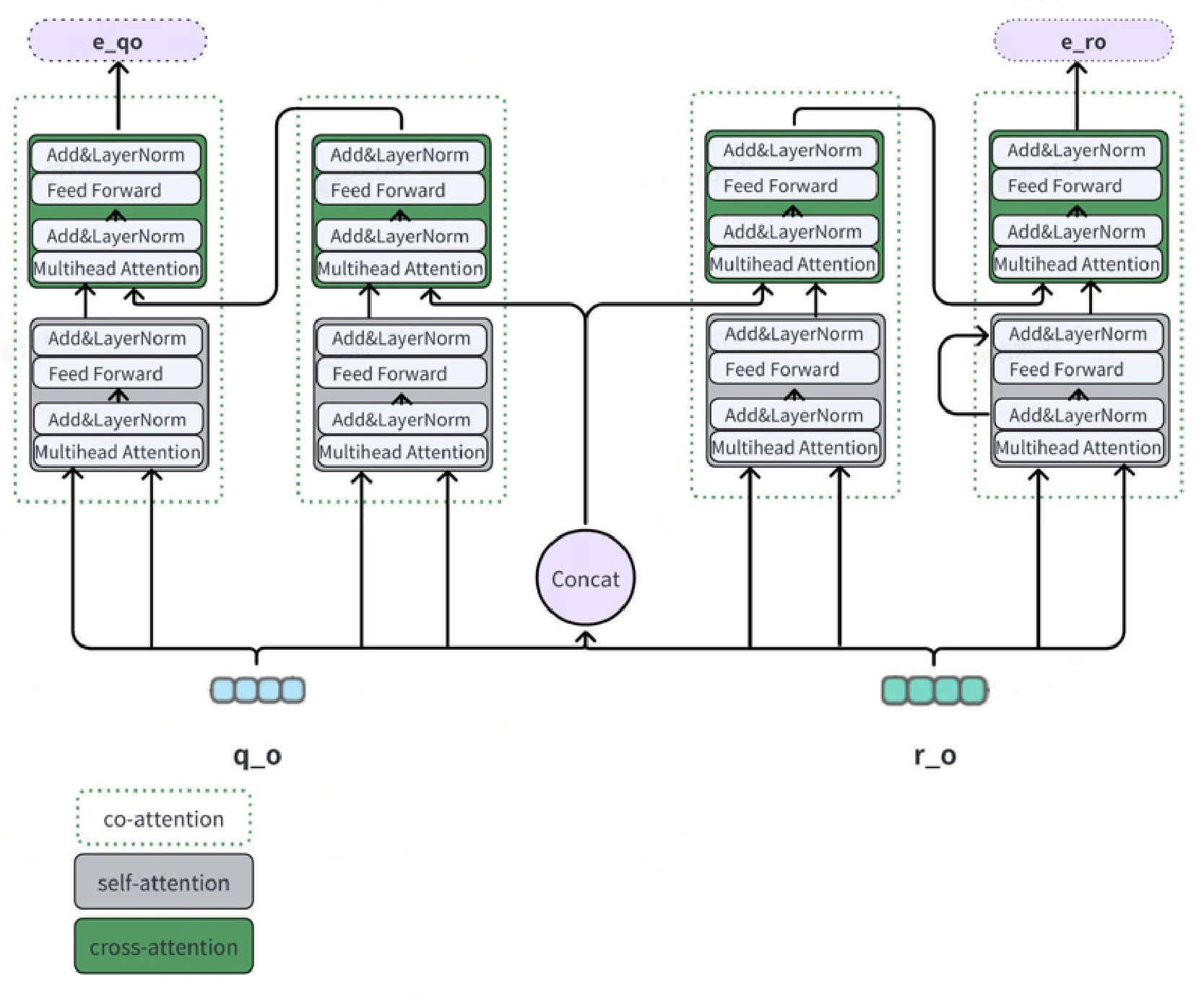}
\end{adjustwidth}
\caption{Co-attention.}
\label{fig:ca}
\end{figure}

To achieve the VCR task, we next design a commonsense encoder layer to capture commonsense between sentences and use it to enhance inference. The~encoder contains $N$ co-attention blocks (Figure~\ref{fig:ca}), each consisting of a cross-attention and a self-attention unit. Its parallel structure enables the model to capture the semantic information between sentences in parallel and ease the information loss problem. A~memory cell follows the $N$-layer co-attention blocks to store the extracted commonsense. We explore each unit in~depth.

\textbf{Self Attention.} Self-attention is designed to capture semantic information within a sequence. Its structure is depicted in Figure~\ref{fig:ca} as a grey block. The~input consists of query $Q$, keys $K$, and~values $V$, which are identical, to~capture pairwise relationships in a sequence. In~more detail, the~multi-head attention layer learns the pairwise relationship between samples in a sequence.
For instance, for~the input sequence $\widetilde{q} = [\widetilde{q_1},\widetilde{q_2},...,\widetilde{q_m}]$, the~multi-head attention learns the relationship between $<\widetilde{q_i}, \widetilde{q_j}>$ and outputs attended representations. Subsequently, the~attended representations are transformed by a feed-forward network which contains two fully-connected layers with ReLU activation and dropout. The~multi-head attention can be formulated as the following:
\begin{equation}
\label{mh1}
MultiHead = h_1 \oplus h_2 \cdots \oplus h_i
\end{equation}
\noindent where $h_i$ represents an attention head and can be expressed as
\begin{equation}
h_i = softmax(\frac{QK^T}{\sqrt{d_k}})V
\end{equation}
\noindent where $d_k$ is the dimension size of input embedding $K$, and~$Q, K, V$ make up the input sequence $\widetilde{q}$.

\vspace{-4pt}

\textbf{Co-Attention.} In comparison to self-attention, co-attention focuses on inter-sentence-wise attention and can be regarded as learning weighted information across different sentences. When taking two different sentence representations, $X = [x_1,x_2,...,x_m]$ and $Y = [y_1,y_2,...,y_m]$, as the inputs, $X$ is the query $Q$ while key $K$ and value $V$ are $Y$, guiding the attention learning process for $X$. Specifically, the~multi-head layer in a co-attention unit attends the pairwise relationship between the two paired input sequences $<x_i, y_j>$ and outputs the attended representations. A~feed-forward layer is then applied to transform the attended representations. The~co-attention network finally outputs $e_{qo}$ and $e_{ro}$, which are attention information over images and~texts.

\textbf{Memory cell.}
Although the commonsense encoder layer improves long-horizon sequence modeling with the attention mechanism, it could have difficulties handling continuous inference, which requires a knowledge base. A~knowledge base, however, is a crucial requirement for cognition-level inference. After~all, even human beings rely on previous knowledge for~inference.

Therefore, we introduce a simple external memory cell to handle this dilemma. A~memory cell is  {readable}--writable 
and learnable. We define the memory cell as $M$, where $M(i)$ has the same size as $e_{qo}$ or $e_{ro}$. At~time step $t$, the~model first reads memory from $M$ regarding past knowledge with the function $f_{read}(M(t-1))$ and then concatenates it with the current embedding $h_t$, which is defined as $f_{write}(M(t))$. This allows the model to condition current embeddings on previous embeddings for a consistent inference. This can be described formally by the following equations:
\begin{equation}
\label{mh2}
f_{read}(M(t-1)) = e_{{qo}/{ro}}^{1} \oplus e_{{qo}/{ro}}^{2} \cdots \oplus e_{{qo}/{ro}}^{t-1}
\end{equation}
\noindent where $\oplus$ represents the concatenate operation.
\begin{equation}
\label{mh3}
m_{q/r} = f_{write}(M(t)) =f_{read}(M(t-1)) \oplus e_{{qo}/{ro}}^{t}
\end{equation}
\noindent where $m_{q/r}$ represents the memory cell output of queries or~responses.

\subsection{Prediction~Layer}
The prediction layer generates a probability distribution of responses from the high-dimension context generated in the encoder layer. It consists of two parts: attention reduction and prediction units, each explored in more detail~below.

\textbf{Attention Reduction.} The commonsense encoder layer includes $N$ layers of co-attention operation. However, some of these are unnecessary for prediction. Therefore, an~attention reduction unit is designed so that only the most significant information is picked up. Mathematically, it can be represented as

\begin{equation}
\label{equ:wt}
\widetilde{Z_l} = \sum_{i=1}^m \alpha_l^{i} z_{l}^i,~\alpha = softmax(MLP(Z_l))
\end{equation}
\noindent where $Z_l$ is either the input query or sequence, $\alpha$ is the learned attention weights, and~$i$ denotes the position in a~sequence.

For better gradient flow through the network, PAVCR also fuses the features by using LayerNorm on the sum of the final attended representations:
\begin{equation}
c =  LayerNorm (W_{x1}^T\widetilde{Z_q} + W_{x2}^T\widetilde{Z_r})
\end{equation}

\noindent where $W_{x1}^T$ and $W_{x2}^T$ are two trainable linear projection~matrices.

\textbf{Prediction.}
The prediction includes a multi-layer perceptron (Dropout(0.3){-}
FC (1024){-}ReLU{-}Dropout(0.3){-}FC(1)). Our VCR task must predict the correct choices from four given choices (answer candidates and reason candidates). Hence, we treat it as a multi-class task.
A popular loss function for multi-classification task {cross-entropy}~ \cite{rubinstein1999cross} 
is therefore applied to complete the prediction. It aims to output probability distributions for four candidate choices, and~the choice with the maximum prediction probability will be the final~prediction.

\section{Experimental~Results}
\label{chap:er}
Here, we conduct extensive experiments to demonstrate the effectiveness of our proposed PAVCR network for solving VCR tasks. We introduce the datasets, evaluation metrics, and~baseline models used for comparison in the experiments. Next, we compare our model to the baseline model, analyze the impact of different techniques, and~present an intuitive interpretation of the predictions. Finally, we conduct ablation studies to investigate our model's performance~further.

The experiments were conducted on a 64-bit machine with a 10-core processor (i9, 3.3~GHz), 64 GB memory, and 3 GTX 1080Ti~GPUs.

\subsection{Dataset}
The VCR dataset~\cite{zellers2019recognition} consists of 290 k multiple-choice questions, corresponding \mbox{290 k} correct answers, 290 k correct rationales, and~110 k images. The~correct answers and rationales are labeled in the dataset with >90\% human agreement. As~referenced earlier in Figure~\ref{fig:running examp}, each set consists of an image, a~question, four available answer choices, and~four reasoning choices. The~correct answer and rationale are provided in the dataset as ground~truths.

\begin{figure}[H]
\includegraphics[width=0.45\textwidth]{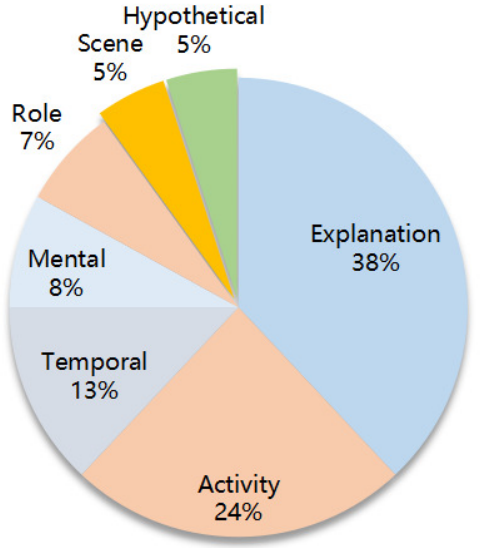}
\caption{Overview of the types of inference required by questions in~VCR.}
\label{fig:Datadis}
\end{figure}

The distribution of the dataset is shown in Figure~\ref{fig:Datadis}. In total, 38\% of the types of inference is about explanation, while 24\% is about activity inference. Hence, these types of tasks represent cognition-level inference~tasks.
\vspace{-4pt}

\unskip

\subsection{Metric}
The VCR task can be regarded as a multi-classification
problem. We use mAP (\textbf{mean average perception})~\cite{henderson2016end} to evaluate performance, a~standard metric for assessing prediction accuracy in multi-classification areas. The~mAP is usually computed on a~dataset.

\subsection{Models for Baseline~Comparisons}
We compare our proposed PAVCR model with recent deep learning-based models for VCR. Specifically, we evaluate against the following baseline approaches:

\begin{itemize}
\item[-] \textbf{RevisitedVQA~\cite{jabri2016revisiting}:}
While the most recent methods proposed for VCR have reasoning modules, RevisitedVQA deploys logistic regressions and multi-layer perceptrons (MLP) for reasoning tasks.
\item[-] \textbf{BottomUpTopDown~\cite{anderson2018bottom}:}
BottomUpTopDown uses weighted feature information over language and images to predict answers and reasons.
\item[-] \textbf{MLB~\cite{kim2016hadamard}:}
The key operation for MLB is the Hadamard product for the attention mechanism, which enables a low-rank bilinear pooling for the task.
\item[-] \textbf{MUTAN~\cite{ben2017mutan}:}
MUTAN consists of a multimodal fusion module tucker decomposition and a multimodal low-rank bilinear (MLB). The~MLB is regarded as a reasoning module for inference.
\item[-] \textbf{R2C~\cite{zellers2019recognition}:}
R2C represents a general baseline for VCR tasks. It consists of a fusion module, a~contextualization module, and~a reasoning module for cognition-level inference. It encodes the sequence based on the sequence relationship model {LSTM}~\cite{hochreiter1997long} and attention mechanism.
\item[-]\textls[-25]{\textbf{DMVCR~\cite{tang2021cognitive}}: In this work, commonsense between sentences is stored using working dynamic memory as a dictionary. In~the inference stage, the~dictionary module looks up information from the dictionary as well as updating the information within the dictionary.}
\item[-]\textbf{CAN~\cite{tang2021interpretable}}: CAN proposes a parallel co-attention-based network to enhance the capability of capturing information from surrounding words.
\end{itemize}

\subsection{Experimental~Results}
\textbf{Task description}
We implemented the experiments in three separate steps. First, we conducted $Q \rightarrow A$ evaluation, followed by $QA \rightarrow R$. Finally, we joined the $Q \rightarrow A$ and $QA \rightarrow R$ results to obtain the final $Q \rightarrow AR$ prediction result. The~difference between the implementation of $Q \rightarrow A$ and $QA \rightarrow R$ tasks is the input query and response. For~the $Q \rightarrow A$ task, the~query is the paired question, image, and~four candidate answers, while the response is the correct answer. For~the $QA \rightarrow R$ task, the~query is the paired question, image, correct answer, and~four candidate rationales, while the response is the correct~rationale.

\textbf{Analysis.}
We compare our method with several popular visual scene understanding models based on the mean average precision (mAP) metric for the three subtasks: $Q \rightarrow A$, $QA \rightarrow R$, and~$Q \rightarrow AR$. As~the results in Table~\ref{tab:experment results} show, our approach outperforms in all subtasks. Specifically, our method outperforms MUTAN and MLB by a large margin, since they lack a commonsense reasoning module, which helps with cognition-level inference. Furthermore, PAVCR also performs better than the recently proposed DMVCR~\cite{tang2021cognitive}.
The reason for this is that the proposed PAVCR model deploys a more efficient multimodal fusion layer and commonsense encoder~layer.

In addition, we compare the model used in the previous work~\cite{tang2021cognitive}, which relies on multiple BiLSTM layers~\cite{huang2015bidirectional}, with~our proposed approach based on multi-head attention modules. This comparison underscores the advantages of our method in addressing the limitations of the BiLSTM-based approach, particularly in handling long-range dependencies in sequential tasks. The~key differences are as follows:

\begin{itemize}
\item Challenges in Capturing Long-Range Dependencies: Although BiLSTMs~\cite{huang2015bidirectional} capture contextual information bidirectionally, their stepwise computation structure inherently limits their ability to effectively model long-range dependencies. Despite using multiple stacked BiLSTMs, these models remain susceptible to vanishing or exploding gradients, especially when processing long sequences. This issue significantly hinders the model's capacity to learn long-term dependencies, representing a critical limitation of the BiLSTM-based method~\cite{tang2021cognitive}.

Our experiments confirm that this limitation directly affects the computational efficiency of BiLSTM-based models, as~they require more time to process long sequences due to the sequential nature of their computations. In~contrast, the~proposed multi-head attention-based approach reduces the inference time by 2.8\% and improves memory usage by 3.1\% (see Table~\ref{tab:efficiency_results}).

\item Advantages of Multi-Head Attention: In contrast, the~multi-head attention mechanism directly models the dependencies between positions in the sequence by computing weighted sums of the sequence elements. This parallelized mechanism eliminates the need for sequential time-step calculations, enabling each attention head to focus on different parts of the sequence. Consequently, it is better equipped to capture long-range dependencies~efficiently.

Unlike BiLSTMs, which struggle with gradient issues over long sequences, multi-head attention overcomes these challenges. It offers a more robust and efficient solution for modeling long-term dependencies, leading to improved computational efficiency. Specifically, PAVCR, which leverages multi-head attention, outperforms the BiLSTM-based DMVCR in terms of both inference time and memory usage (refer to Table~\ref{tab:efficiency_results}).

\item Enhanced Dependency Learning: while BiLSTMs are capable of learning contextual dependencies step by step, their sequential nature restricts their ability to efficiently capture long-range dependencies. In~contrast, the~multi-head attention mechanism processes the entire sequence in parallel, allowing the model to learn dependencies more comprehensively and efficiently across the entire sequence. This parallel processing not only improves the model's ability to handle varying lengths of dependencies, but also significantly boosts performance on long-sequence~tasks.

As a result, the~multi-head attention-based PAVCR demonstrates a substantial reduction in inference time (by 2.8\%) and a noticeable improvement in overall performance (by 3.1\%) compared to the BiLSTM-based approach, as~shown in the experimental results (see Table~\ref{tab:efficiency_results}).
\end{itemize}

This comparison clearly illustrates that the proposed PAVCR model, with~its multi-head attention mechanism, offers significant improvements in both computational efficiency and performance over the BiLSTM-based model in~\cite{tang2021cognitive}. By~addressing the limitations associated with long-range dependencies and computational bottlenecks, our approach provides a more effective solution for the VQA~task.

\begin{table}[H]\setlength{\tabcolsep}{7mm}
\small
\renewcommand\arraystretch{1}
\caption{Overview of our method compared to other popular methods on the VCR dataset using the mean average precision (mAP) metric.
Percentage in parenthesis is our relative improvement over the performance of the best baseline method.}

\vspace{2mm}
\begin{tabular}{l c c c}
\toprule
\textbf{Models} &\boldmath{$Q \rightarrow A$} &\boldmath{$QA \rightarrow R$} &\boldmath{$Q \rightarrow AR$} \\
\midrule

RevisitedVQA~\cite{jabri2016revisiting} & 39.4 & 34.0 & 13.5 \\
\midrule
BottomUpTopDown~\cite{anderson2018bottom} & 42.8 & 25.1 & 10.7 \\
\midrule
MLB~\cite{kim2016hadamard} & 45.5 & 36.1 & 17\\
\midrule
MUTAN~\cite{ben2017mutan} & 44.4 & 32.0 & 14.6\\
\midrule
R2C~\cite{zellers2019recognition} & 61.9 & 62.8 & 39.1\\
\midrule
DMVCR~\cite{tang2021cognitive} & 62.4  & 67.5  & 42.3 \\
\midrule
CAN~\cite{tang2021interpretable} & 71.1 & 73.8 & 47.7\\
\midrule
PAVCR & {73.1 (2.8\%)} 
& 74.2 (0.54\%) & 49.2 (3.1\%) \\

\bottomrule
\end{tabular}
\label{tab:experment results}
\end{table}
\unskip

\begin{table}[H]\setlength{\tabcolsep}{10.7mm}
\small
\renewcommand\arraystretch{1}
\caption{Computational efficiency comparison of PAVCR and other methods on the VCR dataset. The~table shows inference time (in seconds) and memory usage (in GB).}

\vspace{2mm}
\begin{tabular}{l c c c}
\toprule
\textbf{Models} & \textbf{Inference Time (s)} & \textbf{Memory Usage (GB)} \\
\midrule
DMVCR~\cite{tang2021cognitive} & 1.60 & 4.7 \\
\midrule
PAVCR& {1.20} 
& 3.3 \\
\bottomrule
\end{tabular}
\label{tab:efficiency_results}
\end{table}

\textbf{Inference Time (s):} Represents the average time in seconds taken by each model to process a query and return a~result.

\textbf{Memory Usage (GB):} Represents the average memory consumed by each model during inference, measured in~gigabytes.

This is expected as PAVCR incorporates a more effective visual grounding module in its encoder network to enhance visual--textual fusion. In~addition, to~alleviate the information loss suffered by other methods when encoding a long dependence structure for long sentences, PAVCR further encodes semantic information in parallel to capture more comprehensive information from surrounding words, leading to superior performance over the~others.


\subsection{Ablation~Studies}

For thoroughness, we also performed ablation studies to examine the performance of the proposed multimodal fusion layer and the commonsense encoder layer. First, we assessed the impact of the multimodal fusion layer. As~demonstrated in Table~\ref{tab:ab},
removing the multimodal fusion layer leads to a significant drop in results.
This significant decrease in performance indicates that the proposed multimodal fusion layer helps the model improve its capability of capturing visual--textual~information.

\begin{table}[H]\setlength{\tabcolsep}{10.3mm}
\small
\renewcommand\arraystretch{1.2}
\caption{Comparison of results among ablations. Percentage in parenthesis is our relative improvement over the performance of the best baseline~method.}
\vspace{2mm}
\begin{tabular}{l c c}
\toprule
\textbf{Models}  & \boldmath{$QA \rightarrow R$} &\boldmath{$Q \rightarrow AR$} \\
\midrule
\textbf{Without multimodal module} & 40.2 & 38.2 \\
\midrule
\textbf{Dictionary-based encoder} & 63.4 & 69.1 \\
\midrule
\textbf{PAVCR} & {73.1 (15.29\%)} 
&74.2 (7.38\%) \\
\bottomrule
\end{tabular}
\label{tab:ab}
\end{table}

Next, we conducted an ablation study to prove the effectiveness of the proposed commonsense module. On~replacing our encoder layer with a dictionary-based encoder, the~performance decreases
by 13.26\% for the $QA \rightarrow R$ task and by 6.87\% for the $Q \rightarrow AR$ task (see Table~\ref{tab:ab}).
In contrast to the dictionary-based method, the~encoder layer proposed in this work captures information from surrounding words in parallel. This results in less information loss, as~it does not have a long dependency in long sequential prediction~tasks.

\section{Case~Study}
\label{cs}
\begin{figure}[H]
\centering
\includegraphics[height=0.6\textheight, width=\textwidth]{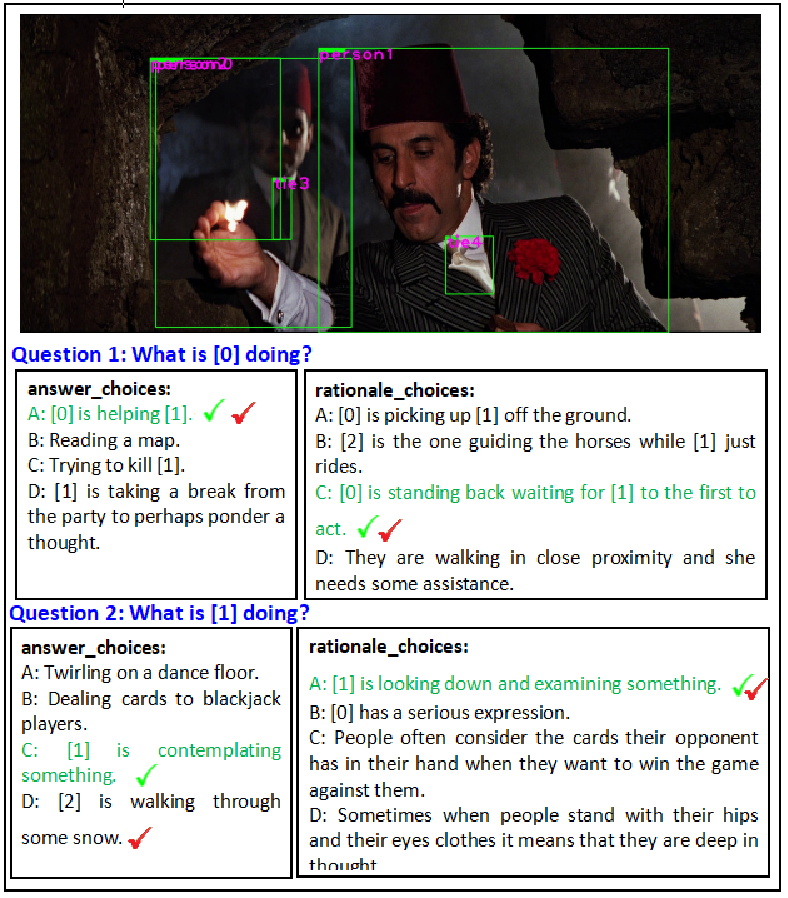}
\caption{Case study example 1. The~model predicts the correct answer and~rationale.}
\label{q4}
\end{figure}

In this section, we present a case study to compare the performance of the proposed method and analyze its superiority.
Figure~\ref{q4} shows prediction results from previous and newly proposed work. The~results of the earlier works are marked in red, while the results from the newly proposed model are in green. The~model predicts two questions for the given image. Both of the two questions regard human activity~prediction.

On analyzing the results in Figure~\ref{q4}, we can conclude that the model proposed in this research performs better than previous work CAN~\cite{tang2021interpretable}, since CAN predicts the wrong answer but correct rationale in Question 2 (see Question 2, Figure~\ref{q4}). In~contrast, our method, PAVCR, can accurately predict each question's correct answer and rationale, demonstrating our model's superior performance. The~multimodal feature fusion layer is crucial to PAVCR's remarkable performance. With~its visual and textual branches, this layer is more powerful in learning visual--textual fusion information, since it considers multimodal fusion while also paying attention to the sequential order, which is a critical factor in semantic~meaning.

\section{Qualitative~Analysis}
\label{qe}

In this section, we conduct a qualitative study of the results from the PAVCR model and present them visually, analyzing how that reflects on our model's cognition-level visual understanding~ability.

\begin{figure}[H]

\includegraphics[height=0.8\textheight,width=\textwidth]{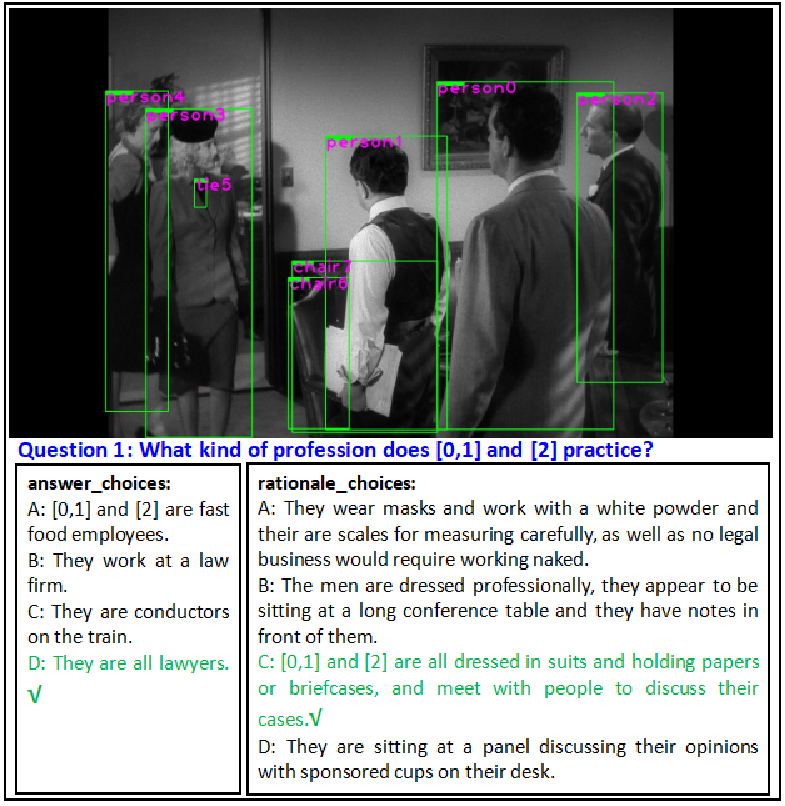}
\caption{Qualitative example 1. The~model predicts the correct answer and~rationale.}
\label{fig:q1}
\end{figure}

\begin{figure}[H]

\includegraphics[height=0.88\textheight,width=\textwidth]{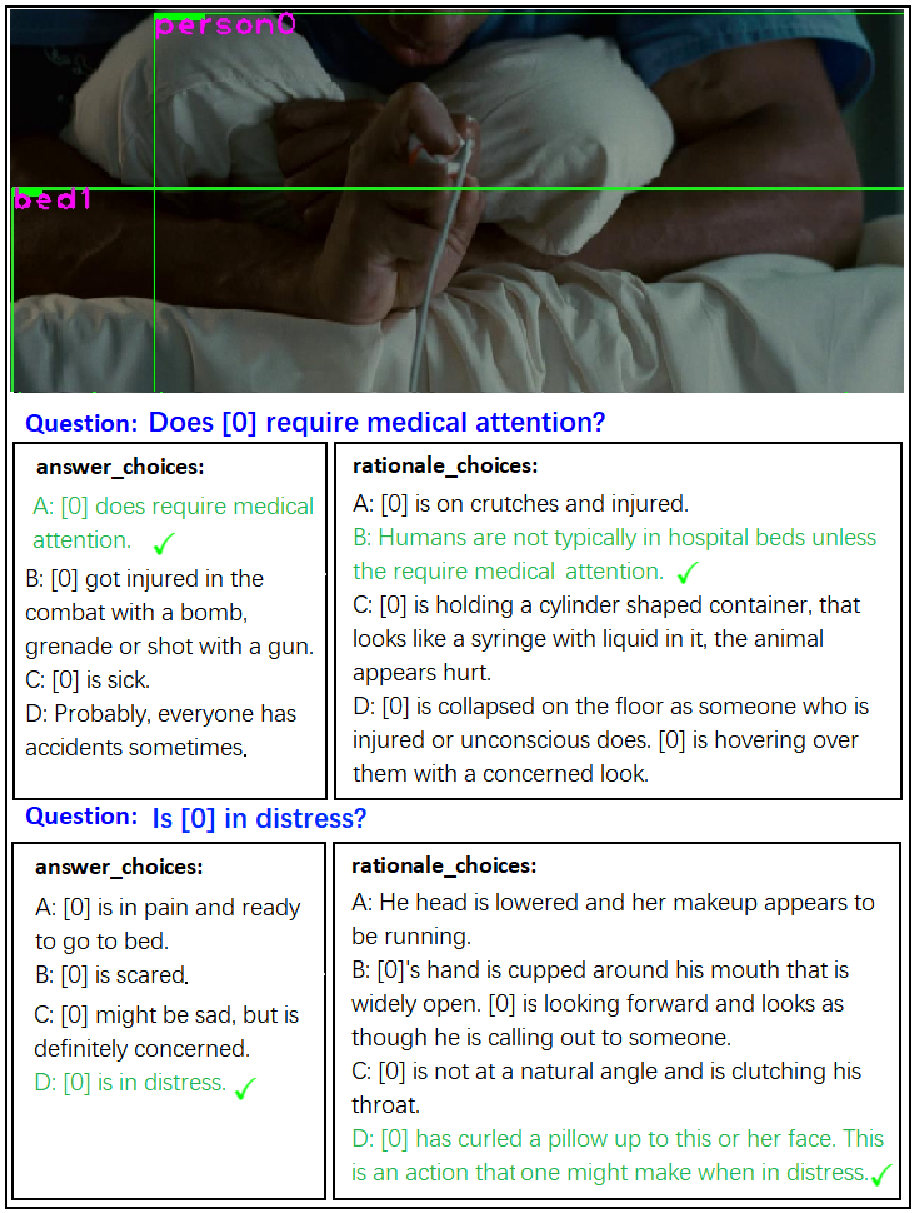}
\caption{Qualitative example 2. The~model predicts the correct answer and~rationale.}
\label{q2}
\end{figure}

\begin{figure}[H]
\centering
\includegraphics[width=\textwidth]{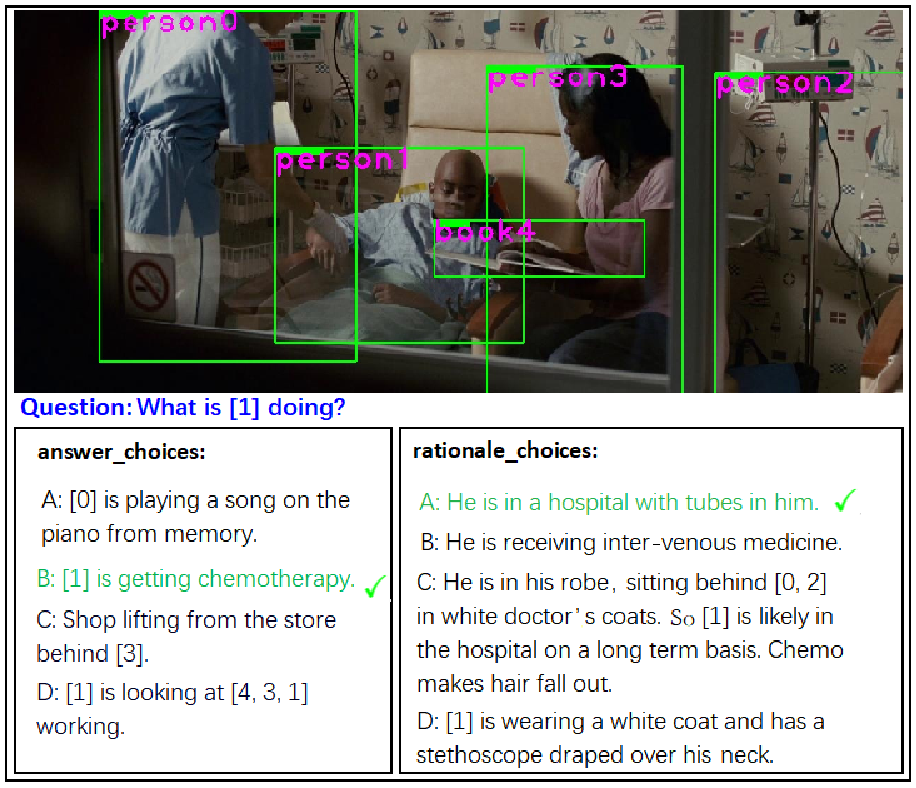}
\caption{Qualitative example 3. The~model predicts the correct answer and~rationale.}
\label{q3}
\end{figure}

The qualitative examples studied are provided in Figures~\ref{fig:q1}--\ref{q3}. We select examples that highlight the varied abilities of our model.
The candidate choice in the green text indicates the correct choice; the candidate with the green checkmark represents the prediction of the PAVCR model.
We selected examples that demonstrate a range of tasks present in the dataset (see Figure~\ref{fig:Datadis}), and~varied application areas. For~example, Figure~\ref{fig:q1} presents an explanation task, while Figure~\ref{q2} depicts an activity and explanation type task. An~activity task is shown in Figure~\ref{q3}. Another motivation for choosing Figures~\ref{q2} and~\ref{q3} is to highlight the performance of our the model on images related to medical settings. We hope to demonstrate the potential contributions this work can have in medical~applications.

In each instance, the~PAVCR model does exceptionally well in both the question-answering and reasoning tasks, highlighting the inference ability of our proposed PAVCR model in cognition-level visual~understanding. 

In Figure~\ref{fig:q1}, the~following question asked is: ``What kind of profession does [0,1] and [2] practice?''. PAVCR can correctly infer the answer and rationale based on elements of dress and activity, even though this question would also be difficult for~humans.

PAVCR can also infer emotion from activities. See, for~example, Figure~\ref{q2}. Question 2 is, ``Is [0] in distress?''. The~model selects the correct answer (``[0] is in distress'') as well as the suitable rationale (``[0] has curled up a pillow to his or her face. This is an action one might make when in distress.'').
\vspace{-6pt}

Finally, the~PAVCR model is capable of predicting human activities based on context. It selects the correct answer (``[1] is getting chemotherapy.'') in Figure~\ref{q3}, along with the right rationale (``He is in a hospital with tubes in him.''), indicating the model can analyze activities based on the surrounding environment. This is an important ability of the model and opens the door for possible real-world~applications.
\vspace{-6pt}

\section{Conclusions}
\label{chap:conclusion}
This work proposes a new model, PAVCR, to~solve the challenging cognition-level visual scene understanding task. With~a novel attention-based multimodal fusion layer and a commonsense encoder layer, PAVCR is able to capture multiple features containing language and object information and fuse them together to maximize efficacy. Extensive experiments, including comparisons with state-of-the-art techniques, ablation studies, and~a case study, demonstrate our model's superior effectiveness and the results' intuitiveness. Future directions include extending the proposed framework in conjunction with our previous work~\cite{tang2021interpretable} to investigate various perspectives of bias in visual~reasoning.

\vspace{6pt}





\authorcontributions{Conceptualization, W.Z.; methodology, W.Z.; software, X.T.; validation, W.Z. and X.T.; formal analysis, X.T.; investigation, X.T.; resources, W.Z.; data curation, X.T.; \mbox{writing---original} draft preparation, X.T.; writing---review and editing, W.Z.; visualization, X.T.; supervision, W.Z.; project administration, W.Z.; funding acquisition, W.Z. All authors have read and agreed to the published version of the~manuscript.}


\funding{{This research received no external funding.} 
}

\institutionalreview{{Not applicable.}
}

\informedconsent{{Not applicable.} 
}

\dataavailability{The original data presented in the study are openly available in the visual commonsense dataset repository at  \url{https://visualcommonsense.com/download/} {(accessed on 20 January 2025).} 
}

\conflictsofinterest{{The authors declare no conflicts of interest.} 
}

\begin{adjustwidth}{-\extralength}{0cm}
\reftitle{References}

\PublishersNote{}
\end{adjustwidth}

\end{document}